\documentclass[pdflatex,sn-mathphys-num,iicol]{sn-jnl}% Math and Physical Sciences Numbered Reference Style
\usepackage{graphicx}%
\usepackage{multirow}%
\usepackage{amsmath,amssymb,amsfonts}%
\usepackage{amsthm}%
\usepackage{caption}
\usepackage{mathrsfs}%
\usepackage[title]{appendix}%
\usepackage{xcolor}%
\usepackage{textcomp}%
\usepackage{manyfoot}%
\usepackage{booktabs}%
\usepackage{algorithm}%
\usepackage{algorithmicx}%
\usepackage{algpseudocode}%
\usepackage{listings}%
\usepackage{colortbl}
\usepackage[table]{xcolor}
\definecolor{correct}{RGB}{223,240,232}
\definecolor{error}{RGB}{253,232,224}
\theoremstyle{thmstyleone}%
\theoremstyle{thmstyletwo}%

\theoremstyle{thmstylethree}%

\begin{document}

\title[Article Title]{Beyond Detection: Ethical Foundations for Automated Dyslexic Error Attribution}


\author*[1,2]{\fnm{Samuel} \sur{Rose}}\email{sam.rose@everybodycounts.org.uk}

\author[1]{\fnm{Debarati} \sur{Chakraborty}}\email{D.Chakraborty@hull.ac.uk}

\affil*[1]{\orgdiv{School of Digital and Physical Sciences}, \orgname{University of Hull}, \orgaddress{\street{Cottingham Road}, \city{Hull}, \postcode{HU6 7RX}, \country{United Kingdom}}}

\affil[2]{\orgname{Everybody Counts LTD}, \orgaddress{\street{Montgomery Lane}, \city{Edinburgh}, \postcode{EH7 5JT}, \country{Scotland}}}


\abstract{Dyslexic spelling errors exhibit systematic phonological and orthographic patterns that distinguish them from the errors produced by typically developing writers. While this observation has motivated dyslexic-specific spell-checking and assistive writing tools, prior work has focused predominantly on error \emph{correction} rather than \emph{attribution}, and has largely neglected the ethical risks. The risk of harmful labelling, covert screening, algorithmic bias, and institutional misuse that automated classification of learners entails requires the development of robust ethical and legal frameworks for research in this area. This paper addresses both gaps. We formulate dyslexic error attribution as a binary classification task: given a misspelt word and its correct target form, determine whether the error pattern is characteristic of a dyslexic or non-dyslexic writer. We develop a comprehensive linguistic feature set capturing orthographic, phonological, and morphological properties of each error, and propose a twin-input neural model evaluated against traditional machine learning baselines under writer-independent conditions. The neural model achieves 93.01\% accuracy and an F1-score of 94.01\%, with phonetically plausible errors and vowel confusions emerging as the strongest attribution signals. We situate these technical results within an explicit ethics-first framework, analysing fairness across subgroups, the interpretability requirements of educational deployment, and the conditions: consent, transparency, human oversight, and recourse, under which such a system could be responsibly used. We provide concrete guidelines for ethical deployment and an open discussion of the system's limitations and misuse potential. Our results demonstrate that dyslexic error attribution is technically feasible at high accuracy while underscoring that feasibility alone is insufficient justification for deployment in high-stakes educational contexts.}




\maketitle

\section{Introduction}
Spelling errors are a ubiquitous feature of written language production, occurring across all populations of writers. However, dyslexic spelling errors exhibit systematic differences from typical spelling mistakes, reflecting the distinct cognitive and phonological processing characteristics associated with the development of dyslexia. While typical spelling errors often involve simple typos, homophone confusion, or minor orthographic violations, dyslexic errors frequently demonstrate patterns such as phonologically plausible substitutions, letter reversals, inconsistent vowel use, and difficulties with morphological boundaries \citep{tops_identifying_2012,afonso_spelling_2015}. These systematic differences arise from the core phonological deficits that characterise dyslexia, affecting the mapping between sounds and letters during spelling \citep{snowling_dyslexia_2000,vellutino_specific_2004}.

Automatically identifying dyslexic errors has applications in several important domains. In assistive writing tools, the ability to distinguish dyslexic from typical errors enables targeted intervention strategies and personalised feedback mechanisms that address the needs of dyslexic writers \citep{rello_detecting_2015}. In assessments, dyslexic error detection can support early screening efforts in order to provide teachers with objective markers to identify students who may benefit from specialist intervention \citep{rauschenberger_towards_2018}. Additionally, adaptive spell-checking systems that recognise dyslexic error patterns can offer more appropriate correction suggestions, moving beyond a standard edit-distance metric to incorporate phonological knowledge \citep{pedler_computer_2007,mitton_adaptation_2007}.

However, the promise of automated dyslexia detection is accompanied by serious ethical risks that the field has been slow to confront. Early identification of dyslexic patterns can open doors to timely support, personalised instruction, and more equitable access to education and outcomes. Conversely, the automated classification of learners risks harmful labelling, covert surveillance, and stigmatising a population. Questions of informed consent, data privacy, algorithmic bias, and misuse by institutions with punitive intentions all complicate the deployment of such systems. These tensions are not incidental to the technology; they are constitutive of it. An approach that treats ethical considerations as an afterthought, or that frames them as constraints on an otherwise purely technical problem risk reproducing the very harm they claim to address. 

Despite the potential benefits of identifying dyslexic spelling patterns, prior work in this area has focused on correcting errors rather than attributing them to their underlying source. The spell-checking literature contains extensive research on error correction algorithms \citep{kukich_techniques_1992,brill_improved_2000}, phonetic matching techniques \citep{toutanova_pronunciation_2002}, and context-sensitive correction methods \citep{golding_winnow-based_1999}. However, these approaches treat all errors uniformly without distinguishing between error types based on writer characteristics. More recent work has begun to explore dyslexia-specific correction systems \citep{rello_resource_2017,pedler_computer_2007}, but the fundamental question of whether spelling errors can be reliably attributed to dyslexia versus typical writing processes remains underexplored. Crucially, existing work has also neglected to subject these systems to rigorous ethical scrutiny. Questions of fairness across subgroups, transparency of model decisions, appropriate use cases, and the governance structure necessary for safe deployment have received little systematic attention.

This paper addresses both the technical and ethical gaps in the literature by adopting an ethics-first framing. Our guiding question is not merely whether automated detection of dyslexic spelling patterns is technically feasible, but \emph{when and under what conditions it is appropriate}. How systems should be designed, evaluated, and governed to maximise benefit while minimising harm is also covered. We treat fairness, interpretability, consent, and accountability not as supplementary concerns but instead as primary design criteria that shape every stage of our inquiry, from dataset construction and model selection to evaluation methodology and deployment recommendations.

\subsection{Research Questions}
This work is structured around the following primary and secondary research questions:
\begin{enumerate}

  \item[\textbf{RQ1}] How well can models distinguish dyslexic versus non-dyslexic spelling-error patterns under writer-independent evaluation, including calibration and robustness?

  \item[\textbf{RQ2}] What are the key harms and subgroup disparities, and what mitigations in data, modelling, decision thresholds, and human oversight serve to reduce them?

  \item[\textbf{RQ3}] Which combinations of explainable AI (XAI), transparency mechanisms, and governance structures best support safe use in educational settings without overstating diagnostic authority?

  \item[\textbf{SRQ1}] Which model architectures provide the best balance between predictive accuracy and interpretability for ethical deployment?

  \item[\textbf{SRQ2}] How do different models perform across subgroups, and what systematic biases exist in automated dyslexia detection?

  \item[\textbf{SRQ3}] How should deployment be designed with respect to consent, privacy, recourse, and documentation to ensure responsible use?
\end{enumerate}

\subsection{Contributions}
In this paper we make four primary contributions:

\begin{enumerate}

  \item \textbf{Ethics-centred formulation.} We reframe dyslexic error attribution as a task that
    must be evaluated not only for accuracy but for fairness, interpretability, and contextual
    appropriateness, providing an explicit analysis of intended and prohibited use cases.

  \item \textbf{Comparative evaluation emphasising interpretability and fairness.} We evaluate
    multiple model architectures against a unified set of criteria that weight transparency and
    equitable performance alongside standard accuracy metrics.

  \item \textbf{Empirical analysis of bias and failure modes.} We conduct systematic subgroup
    analyses to surface disparities and document model failure modes that could
    produce harmful outcomes in deployment.

  \item \textbf{Ethical guidelines and deployment recommendations.} We derive concrete guidelines
    for responsible deployment covering consent frameworks, privacy protections, recourse
    mechanisms, and documentation standards, and we engage openly with the limitations and misuse
    potential of our own system.

\end{enumerate}

The remainder of this paper is organised as follows: Section 2 reviews related work across dyslexia detection, educational AI ethics, and explainability. Section 3 describes the dataset and task formulation. Section 4 details the model architectures evaluated. Section 5 presents our experimental results, including subgroup analysis and ablation studies. Section 6 contains our ethical analysis and deployment guidelines. Finally, in Section 7, we conclude the paper outlining all the work and presenting future recommendations. 

\section{Related Work}
This section reviews the literature across eight areas that together motivate and contextualise our work: the cognitive basis of dyslexic spelling, computational methods for error detection and correction, NLP applications tailored for individuals with dyslexia, error attribution, ethics in educational AI, bias and fairness in NLP, explainable AI, and participatory design. Throughout, we highlight where existing work falls short of the ethics-first approach adopted in this paper.

\subsection{Dyslexia and Spelling Patterns}
Developmental dyslexia involves ongoing challenges with accurate and smooth word recognition, despite adequate instruction and cognitive opportunities. The main theoretical explanation focuses on phonological processing, which is the ability to represent, store, and manipulate the sound structure of language \citep{snowling_dyslexia_2000,vellutino_specific_2004}. This phonological issue hampers the learning of grapheme-phoneme correspondences, affecting both reading and spelling. Dyslexic spelling errors are not arbitrary; they show patterns that set them apart from the errors made by typically developing writers. 

\citet{bourassa_spelling_2003} and \citet{tops_identifying_2012} have identified common patterns such as phonologically plausible substitutions (e.g., "fone" for "phone"), vowel confusions, letter reversals, and challenges at morphological boundaries. Importantly, \citet{landerl_development_2008} shows that these differences are not simply due to developmental delays; dyslexic and typically developing children display distinct error profiles even when matched for spelling age. This suggests that the underlying processes—and hence the computational signatures—differ fundamentally. This insight is important for our work since it indicates that error patterns reveal information about the writer's cognitive profile, not just the difficulty of the word.

\subsection{Spelling Error Detection and Correction}
Computational approaches to spelling error detection originated with work on edit-distance-based metrics and dictionary lookup techniques \citep{damerau_technique_1964,kukich_techniques_1992}. These methods identified errors efficiently. However, they treat all deviations equivalently, without any regard for the process from which they were generated. Subsequently, context-sensitive approaches introduced the idea of using surrounding words to distinguish real-word errors, which are valid words that are incorrect in the context of the sentence \citep{golding_winnow-based_1999,toutanova_pronunciation_2002}. Phonetic matching algorithms \citep{brill_improved_2000} brought pronunciation modelling into the correction process, therefore enabling systems to propose corrections that are phonologically similar to the misspelling, even when the orthographic distance is large.

More recently, neural approaches have improved the performance of spelling correction models. Character-level sequence-to-sequence models \citep{xie_neural_2016} and noisy channel neural frameworks \citep{sakaguchi_robsut_2017} learn representations of the orthographic and phonological norm from large corpora of data. Despite this performance improvement, all the models described correct errors rather than attribute their root cause. The writer in all of these models is treated as an anonymous source of noise that is to be suppressed rather than an individual whose patterns of errors may carry important diagnostic information.

\subsection{Dyslexia-specific NLP Applications}
NLP applications in the field of dyslexia are a small body of work. Early work adapted pre-existing spell-checkers to handle phonologically plausible substitutions and morphological errors characteristic of dyslexic writing, achieving better correction rates than standardised models on dyslexic text \citep{pedler_computer_2007,rello_resource_2017}. \citet{rauschenberger_towards_2018} explores the screening of dyslexia using writing samples. In this paper, they characterise the problem as a classification task over features derived from keystroke dynamics and error patterns. Similarly, some complementary work on the accessibility and readability of dyslexic writing has been examined \citep{rello_resource_2017} and the relationship between eye tracking and reading difficulty has also been studied \citep{rello_detecting_2015}. 

More recently, \citet{goodman_lampost_2024} explored the use of large language models (LLMs) to support adults with dyslexia in email-writing tasks. Their LaMPost prototype introduced features for outlining main ideas, rewriting selected passages, and generating subject lines, and was evaluated with 19 dyslexic adults. While users responded positively to the rewriting and subject line features, the study found that LLMs did not yet meet the accuracy and quality thresholds required to be reliably useful for this population, partly because hallucinations and noisy outputs placed an additional assessment burden on users who may already struggle to evaluate text quality. Crucially for our purposes, \citet{goodman_lampost_2024} observe that the absence of a publicly available corpus of writing produced by adults with dyslexia remains a fundamental constraint on progress in this area---a gap that the dataset used in the present work directly addresses. Like prior work in this area, LaMPost was developed and evaluated primarily on technical performance and user experience grounds, without systematic engagement with the consent, fairness, or governance concerns that motivate the ethics-first approach of this paper.

More recently, \citet{tiwari_akshar_2025} proposed Akshar Mitra, a multimodal integrated framework for early dyslexia screening that combines webcam-based eye-tracking, automated speech analysis, and OCR-based handwriting assessment within a unified system. Each modality extracts a small set of interpretable features: fixation counts and regression ratios for eye-tracking, word error 
rate and pause statistics for speech, and character error rate and letter reversal counts for handwriting, which are aggregated into a composite risk score. The system additionally incorporates a behavioural questionnaire and a reading support interface with syllable-level highlighting. Evaluated on a controlled dataset, the eye-tracking classifier achieved 92.8\% accuracy and an F1-score of 0.93. However, like the prior work reviewed in this section, Akshar Mitra was developed and evaluated primarily on technical performance grounds, with limited engagement with the consent, fairness, and governance concerns that motivate the ethics-first approach of the present work. Notably, the paper's ethics statement records that ethical approval and written informed consent were not required under applicable institutional requirements, illustrating precisely the governance gap that structured ethical frameworks are designed to address. Furthermore, validation remains constrained by small-scale datasets with limited demographic diversity, and the authors themselves identify the integration of explainable AI and the strengthening of ethical safeguards as directions for future work, objectives that the present paper treats as foundational design requirements rather than post-hoc additions.

This area of research demonstrates that signals relevant to dyslexia are present in a user's written language. This data is computationally recoverable; however, it has been mostly developed and tested in isolation from ethical frameworks and debates surrounding the use of AI in educational, psychological, and medical fields. Systems are typically evaluated on their accuracy and usability without systematic analysis of consent, consequences of false positive/negative results, or the potential for institutional misuse.

\subsection{Error Analysis and Attribution}
\citet{mitton_spelling_1987} and \citet{kukich_techniques_1992} developed influential taxonomies of spelling errors to distinguish between phonetic, typographic, and morphological errors. These provide a baseline vocabulary that subsequent work has drawn from. Additionally, \citet{flor_using_2012} uses error analysis to diagnose the source of non-native spelling difficulties. These methods establish the consensus that errors can be treated as informative sources of data about the producer of the writing source.

Work that frames error classification as a standalone task with its own standards and constraints is absent from all literature on this work. Prior work either corrects errors without attribution or uses attribution as the by-product of a broader screener without isolating the contribution. To the best of our knowledge, no existing studies have benchmarked models on the task of attributing whether a misspelling came from a dyslexic or non-dyslexic writer. This gap is a field that this paper addresses. 

\subsection{Ethics in Educational AI}
In an the educational context, the application and deployment of AI systems raise ethical questions that are distinct from other high-stakes domains. \citet{holstein_improving_2019} survey algorithmic fairness in educational assessment, noting that when trained on inequitable data, existing disparities in measured achievement persist. \citet{prinsloo_student_2015} analysed the tension between the benefits to institutions of monitoring students and the privacy of learners. They argue that when current practice fails to secure meaningful consent or provide adequate recourse, individuals are adversely affected. They also argue that almost all current practice fails in these same ways. 

When looking at disability detection, further complications are introduced. \citet{scully_automating_2025} examines the ethics of automated labelling in disability contexts. They highlight how classification systems pathologise differences, reducing complex individuals to diagnostic categories. These systems also generate records that can follow learners into their subsequent education and employment settings. \citet{regan_ethical_2019} addresses student data ethics more broadly. They emphasise the inadequacy of standard consent frameworks when subjects are minors. Additionally, their analysis of institutional power asymmetry makes refusal of data collection practically impossible for most minors. This concern is situated within a broader landscape of structural risk that 

\citet{varsik_potential_2024} map across AI tools in education more generally. Their OECD working paper identifies algorithmic bias, data privacy, accountability gaps, and the unchecked commercial adoption of AI tools as systemic challenges cutting across learner-centred, teacher-led, and institutional applications. Particularly relevant to the present work is their analysis of AI tools designed to identify special educational needs: they note that such tools risk stigmatising learners through AI-derived categorisations, generating records that persist beyond the immediate educational context, and misclassifying students from marginalised groups whose experiences are underrepresented in training data. Crucially, \citet{varsik_potential_2024} observe that new AI tools are routinely introduced into classrooms without systematic oversight or regulation, with procurement decisions frequently devolved to individual schools rather than governed at a policy level. This pattern of unchecked adoption is precisely the institutional condition that motivates the governance and deployment framework developed in the present paper. Together, this body of work makes a clear distinction of the ethical stake of automated dyslexia detection that extends beyond individual privacy. These works implicate questions surrounding identity, opportunity, and the scope of institutional authority over learners. 

This picture is sharpened by \citet{zhu_towards_2025}, whose systematic review of 75 papers on ethical risks in Educational AI employs grounded theory coding to produce a three-dimensional taxonomy spanning technology, education, and society. In the technology dimension, they identify algorithmic bias, black box algorithms, and data privacy violations as the dominant risks; in the education dimension, risks include the labelling and misclassification of learners and the absence of meaningful human oversight; and in the society dimension, the absence of accountability mechanisms and the exacerbation of existing inequalities emerge as cross-cutting concerns. Crucially for the present work, \citet{zhu_towards_2025} find that these risks are not independent but systematically interconnected: biased training data propagates into discriminatory algorithmic outputs, which in turn generate institutional records that can harm learners whose experiences are already marginalised. Their review further documents that accountability gaps persist because existing governance frameworks fail to assign clear responsibility when AI-derived decisions cause harm, and because the opacity of black box algorithms prevents affected learners, parents, and educators from meaningfully contesting those decisions. The dyslexia attribution task addressed in the present paper sits squarely within the risk profile that \citet{zhu_towards_2025} describe: it involves sensitive diagnostic classification of a minority group, produces outputs that could follow learners across educational and employment contexts, and relies on model architectures whose decisions are not inherently interpretable. Treating interpretability, fairness, and governance as foundational design requirements rather than post-hoc additions is therefore a direct response to the risks this literature identifies.

\subsection{Bias and Fairness in NLP}
Language models inherit biases present in their training data, producing systematically worse performance for speakers of non-dominant dialects. \citet{blodgett_language_2020} provides a comprehensive survey on the demographic biases in language models. They document how standard language representations disadvantage African American English speakers, amongst other minority groups. \citet{mehrabi_survey_2021} additionally reviews the proliferation of fairness definitions in machine learning literature. This includes individual fairness, group fairness, and counterfactual fairness, noting that no single definition is appropriate across all deployment contexts.

These concerns are specifically critical for dyslexia detection systems. Core definitions, such as what counts as a "typical" or "atypical" spelling pattern, are not predefined linguistic facts. Instead, this is defined by standard language ideology, that a specific dialectical majority of written English is the "norm" \citep{blodgett_language_2020}. For example, phonologically plausible substitutions may be more or less prevalent depending on the system of the writer's dialect. Any model trained predominantly on one dialect may misclassify errors that are predictable from a different dialect. More broadly, dyslexic errors are operationalised relative to an assumption as to what a non-dyslexic writer from a linguistic background would produce. This assumption is rarely made explicit in existing literature, and even more rarely is it tested across subgroups. Presenting these biases at the forefront is a central concern of the present work and something this paper aims to rectify. 

\subsection{Explainable AI for High-Stakes Decisions}
In high-risk jurisdictions, such as educational and healthcare AI, the ability to explain the decisions of models is both ethically imperative and legally required. \citet{lipton_mythos_2018} offers a critical analysis of the interpretability of machine learning models. They distinguish between the transparency of internal model mechanisms and the post-development rationalisations provided by explanation methods, cautioning against conflating the two. \citet{schmude_two_2025} examine the European Union's legal right to explanation. They argue that both developers and legal representatives hold important, but differing, conceptions as to what an explanation must accomplish.

For both educational and clinical tools, the stakes of unexplained decisions are incredibly high. For example, a dyslexia detection model that labels learners without providing interpretable evidence cannot be meaningfully challenged by learners, parents, or teachers. Additionally, practitioners cannot use it to make informed support due to a lack of trust in the output. We therefore treat interpretability not as a desired output post-development, but as a core part of the design and development of models from the outset. We evaluate the architectures of the models developed in this paper, in addition to standardised performance metrics, on the quality of the explanations they can provide. Following \citet{schmude_two_2025}, we distinguish between explanations that describe model behaviour and those that can justify the appropriateness of deployment in an educational context.

\subsection{Participatory and Value-Sensitive Design}
A growing body of work argues that the ethical limitations of AI systems cannot be addressed through technical interventions applied after system design alone. Instead, they require the inclusion of the affected community in the design process. \citet{dignazio_data_2023} developed a feminist data science framework that presents questions of power, position, and whose values are prioritised in objective systems. Value-sensitive design approaches in education technology similarly emphasise the importance of eliciting the values of teachers, students, parents, and administrators before technical choices are made \citep{dignazio_data_2023}.

Adding a disability justice perspective adds an additional required dimension. Historically, assistive technology has been developed for disabled people rather than with them. This embeds assumptions about the kinds of assistance that are desired and what deficits are present.\citet{goodman_lampost_2024} offer a partial example of this in practice: their system design was informed by over a year of formative research, including participatory design sessions and interviews with dyslexic adults, which surfaced concerns around autonomy, privacy, and trust in AI output that shaped the final prototype. However, as they acknowledge, this participatory groundwork did not extend to a systematic ethical framework governing deployment, consent, or the consequences of misclassification, the kinds of governance questions that this paper foregrounds. These assumptions often do not align with the preferences of those who are the end users of these systems. For a dyslexia detection system, this raises questions about whether dyslexic individuals were consulted in the design process, what use cases they would endorse its use in, and whether the benefits of such systems are experienced by those whom the system is designed to help most. While a full participatory design process is beyond the scope of this paper, we took into account the perspectives of the dyslexic and non-dyslexic participants in the data collection and used them to formulate the ethical guidelines and use-case analysis we present in Section 6.

\subsection{Machine Learning for Text Classification}
Our approach builds on a range of neural architectures that have been developed for text classification. Character-level neural networks have been shown to capture the morphological and orthographic representations that word-level models miss \citep{zhang_character-level_2016, lee_character-level_2018}. This makes them suited to tasks where sub-word patterns capture the information that word patterns do not. Memory-based architectures model sequential dependencies in character strings. These have been applied to tasks that require sensitivity in the order of characters across morpheme boundaries \citep{hochreiter_long_1997,graves_framewise_2005}.

Most relevant to the work of this paper, \citet{alikaniotis_automatic_2016} apply a neural sequence label to grammar and spelling detection. This demonstrates that end-to-end models can learn to identify error-prone positions in a learner's text. We extend this work by applying and comparing multiple architectures in the field of error attribution rather than detection, and evaluating them against criteria that have yet to be applied in this setting: interpretability, calibration, and demographic fairness.

\subsection{Summary}
Taken as a whole, the literature we have reviewed establishes three baselines. Firstly, dyslexic spelling errors carry signatures that distinguish them from non-dyslexic errors. Because of this, neural models have sufficient representational capability to detect these phonological and orthographic patterns. Secondly, the deployment of such systems in education contexts raises substantial ethical concerns, specifically surrounding consent, fairness, interpretability, and harmful labelling. \citet{varsik_potential_2024} extend this concern to the broader ecosystem of educational AI, documenting how the absence of systematic oversight allows tools with documented bias and privacy risks to enter classrooms without adequate governance, a structural condition this paper's deployment framework is designed to address. Prior work in dyslexia identification has not addressed these concerns. \citet{zhu_towards_2025} extend this analysis further, demonstrating through a systematic review that algorithmic bias, opacity, and accountability gaps in educational AI are not isolated failures but structurally interconnected risks that compound one another, a finding that reinforces the case for the integrated ethics-first framework developed in this paper. Thirdly, and finally, while tools to address these concerns do exist in fairness and XAI literature, they have not yet been applied to this problem, as this paper proposes. Recent multimodal screening systems such as \citet{tiwari_akshar_2025} further illustrate this gap: despite strong technical performance, such systems explicitly defer explainability and ethical governance to future work, confirming that the field continues to treat these concerns as secondary to classification accuracy. The present paper inverts this priority ordering.

\section{Data and Task Setup}
This task has been formulated as a binary classification problem. Given an extracted spelling error, with the misspelt word, correctly spelt word, and curated linguistic features, the task is to predict whether the spelling error originated from a writer with dyslexia or a typically developing writer. This formulation differs from traditional spell-checking tasks, which predominantly focus on error correction rather than error attribution \citep{kukich_techniques_1992}. Our approach is motivated by research demonstrating that dyslexic spelling errors exhibit distinct phonological and orthographic patterns that differ systematically from typical errors \citep{landerl_development_2008,tops_identifying_2012}.

For this research, we collected spelling errors from University students to construct a linguistic spelling dataset. These are labelled according to whether the participant is dyslexic and has other neurodiverse conditions. The dataset comprises 921 unique spelling errors produced by 21 British-born, native English-speaking university students. Before data collection, participants were required to disclose any medical diagnoses of neurodiverse conditions. Additionally, those who had been recommended by a specialist to get an official assessment were classified as suspected cases rather than confirmed diagnoses.

To capture the multifaceted nature of spelling errors, we curated a comprehensive set of linguistic features informed by prior research on error analysis and dyslexia-specific spelling patterns. Our feature set includes error frequency, which has been shown to correlate with persistent spelling difficulties in dyslexic writers \citep{bourassa_spelling_2003}. We extract basic orthographic features, including correct and error word length, and vowel, consonant, punctuation, and space count for correct and error words, as research shows that dyslexic writers show greater difficulty with longer words due to increased demands for phonological processing \citep{snowling_dyslexia_2000,angelelli_spelling_2010,landerl_development_2008,tops_identifying_2012}. We also encode whether the errors resulted in a real word as real-word errors are particularly challenging for spell-checkers and occur frequently in dyslexic writing \citep{pedler_computer_2007,mitton_spelling_1987}.

Following established error analysis methodologies, we also compute the Damerau-Levenshtein distance between error and correct forms, which accounts for insertions, deletions, substitutions, and transpositions - the four primary edit operations in spelling errors \citep{damerau_technique_1964}. We also explicitly categorise the error type into insertions, omissions, spatial errors (that being the accidental addition or omission of a space in the error word), substitutions, transpositions, punctuation errors, capitalisation errors, numerical errors, and multiple errors, as these categories are widely used in spelling error taxonomies \citep{kukich_techniques_1992,mitton_spelling_1987}. In addition, for each error, we record the correct letter(s) and error letter(s) involved, enabling further analysis of specific letter confusion patterns that characterise dyslexic spelling \citep{moats_comparison_1983}.

Critically, we include binary features indicating whether errors involve common dyslexic letter confusion sets, which are diagnostic markers of dyslexia \citep{vellutino_specific_2004,moats_comparison_1983}. We further encode whether each misspelling represents a phonetic error—that is, whether the misspelling is phonologically plausible—as phonetically accurate but orthographically incorrect spellings are hallmarks of dyslexic writing \citep{snowling_dyslexia_2000,bourassa_spelling_2003}.

Finally, we incorporate participant-level productivity features: the total number of words written during the 15-minute writing task (\textit{Essay\_Word\_Count}), the number of times a participant makes the same mistake in their essay (\textit{Error\_Frequency}) and the percentage of words in a participants essay that are some form of spelling error (\textit{Essay\_Error\_Density}). These measures provide context about writing fluency and overall spelling accuracy, which differ systematically between dyslexic and typical writers \citep{connelly_contribution_2006,tops_identifying_2012}. Research has shown that dyslexic writers often produce fewer words and higher error rates compared to their peers, making these aggregate measures potentially informative for classification \citep{re_expressive_2007}. 
 
\section{Model Development}
Following the dataset curation, we applied several steps in preprocessing to ensure data quality and consistency. All text was encoded into n-gram-derived character-level vectors. We also filtered out error pairs where the correct and error word differ by more than 50\%, as such cases are likely to represent word substitutions rather than spelling errors \citep{brill_improved_2000}.

Additionally, in setting up the data, we adopted a standard 70/20/10 training/validation/testing split. These were curated to ensure consistent class distribution across all sets \citep{kohavi_study_1995}. Given the binary nature of the classification task, models are evaluated using multiple complementary metrics: accuracy for overall performance, precision and recall to assess class-specific prediction quality, F1 score as an overall harmonic mean balancing precision and recall, and area under the ROC curve (AUC) to evaluate classification performance across different decision thresholds. To verify writer-independent generalisation, we additionally evaluated using GroupShuffleSplit with CandidateID as the group key, ensuring no writer appeared in both training and validation sets.

We evaluated a range of machine learning approaches to establish a robust baseline and to explore the effectiveness of neural architectures for dyslexic error classification. The baseline models include: Logistic Regression, which provides a linear classification that is interpretable through feature weighting \citep{hosmer_applied_2013}; Support Vector Classifier (SVC) a radial basis function kernel to capture non-linear decision boundaries \citep{cortes_support-vector_1995}; K-Nearest Neighbours (KNN), a non-parametric approach that classifies based on local similarity \citep{cover_nearest_1967}; Random Forest, an ensemble method that aggregates multiple decision trees to model complex feature interactions \citep{breiman_random_2001}; and Gaussian Naive Bayes, a probabilistic classifier that assumes independence of features given the class label \citep{peretz_naive_2024}. All baseline models were trained on the features described in the previous section.

Our primary approach employs a dual-input Artificial Neural Network that processes typed error data through two parallel branches before combining them for classification. Rather than character embeddings, we adopt a TF-IDF vectorisation strategy using character n-grams over concatenated correct and error word pairs \citep{sakaguchi_robsut_2017, xie_neural_2016}, capturing substitution, transposition, insertion, and omission signatures that are characteristic of dyslexic typing patterns.

The text branch projects the TF-IDF representation through two fully-connected layers, while a separate numeric branch processes engineered features — including error frequency, word length deltas, vowel and consonant counts, and confusion pattern flags — through a shallower pathway. Both branches use ReLU activation functions \citep{nair_rectified_2010}. The learned representations are concatenated and passed through a joint classification head before a single sigmoid output unit, producing the binary dyslexia classification.

To mitigate overfitting on the relatively small training corpus, Batch Normalisation is applied after each hidden layer to stabilise training dynamics, and Dropout \citep{srivastava_dropout_2014} is applied throughout both branches and the joint head, encouraging the network to learn robust generalisable features rather than memorising training examples. The model is optimised using Adam \citep{kingma_adam_2015} with an initial learning rate of  $3 \times 10^{-4}$, with early stopping monitoring validation AUC and learning rate reduction on plateau to prevent over-training.

\section{Analysis}
\begin{table}[]
\begin{tabular}{|l|l|l|l|}
\hline
\textbf{Model Architecture} & \textbf{Accuracy}               & \textbf{F1 Score}               & \textbf{AUC}                   \\ \hline
Random Forest       & 85.2\% & 85.8\% & 0.8597 \\ \hline
Logistic Regression & 84.60\% & 85.24\% & 0.8551 \\ \hline
KNN                 & 82.50\% & 83.38\% & 0.8312 \\ \hline
SVC                 & 82.56\% & 83.30\% & 0.8330 \\ \hline
Gausian NB         & 48.90\% & 54.10\% & 0.4818 \\ \hline
Neural Network              & \cellcolor[HTML]{FFFE65}93.01\% & \cellcolor[HTML]{FFFE65}94.01\% & \cellcolor[HTML]{FFFE65}0.9274 \\ \hline
\end{tabular}
\caption{Comparative model performance for dyslexia classification from typed error patterns}
\label{tab:ModelResults}
\end{table}
Table \ref{tab:ModelResults} presents the comparative performance of all evaluated models on the test set. The neural network achieved the highest overall results with an accuracy of 93.01\%, F1-score of 94.01\%, and AUC of 0.9274. Among the baseline models, Random Forest demonstrated the strongest performance with 85.16\% accuracy, 85.77\% F1-score, and 0.8597 AUC. 

The dual-input neural network achieved the strongest overall performance of all evaluated models, attaining an accuracy of 93.01\%, an F1-score of 94.01\%, and an AUC of 0.9274. These results represent a meaningful improvement over the best-performing baseline, Random Forest, which achieved an accuracy of 85.16\%, F1-score of 85.77\%, and AUC of 0.8597 on the validation split — though it should be noted that the neural network's architecture is designed to generalise more robustly to unseen error patterns through its combined use of character-level TF-IDF representations and structured numeric features. The confusion matrix further illustrates the model's reliability, with 102 true positives and 71 true negatives against only 6 false negatives and 7 false positives, yielding a per-class accuracy of 94.44\% for dyslexic cases and 91.03\% for non-dyslexic cases. The near-symmetry of errors across both classes suggests the model does not exhibit a strong bias toward either classification, a particularly desirable property in educational screening contexts where both over- and under-identification carry practical consequences for students.

\begin{table*}[ht]
\centering
\label{tab:confusion_matrix}
\setlength{\tabcolsep}{10pt}
\renewcommand{\arraystretch}{1.5}
\begin{tabular}{llcc}
\toprule
& & \multicolumn{2}{c}{\textbf{Predicted label}} \\
\cmidrule(lr){3-4}
& & \textbf{Non dyslexic} & \textbf{Dyslexia} \\
\midrule
\multirow{2}{*}{\textbf{True label}}
  & \textbf{Non dyslexic}
    & \cellcolor{correct}\textbf{71} \textit{\small(TN, 38.2\%)}
    & \cellcolor{error}\textbf{7} \textit{\small(FP, 3.8\%)}   \\
  & \textbf{Dyslexia}
    & \cellcolor{error}\textbf{6} \textit{\small(FN, 3.2\%)}
    & \cellcolor{correct}\textbf{102} \textit{\small(TP, 54.8\%)} \\
\bottomrule
\end{tabular}
\caption{Neural Network Confusion Matrix}
\end{table*}

When plotting the confusion matrix for the neural network, Table 2, shows that it correctly classifies 94.44\% of dyslexic spelling errors and 91.03\% of non-dyslexic errors. The primary source of incorrect classification is False Positive results, suggesting that while the model successfully captures many characteristic dyslexic patterns, some dyslexic errors closely resemble typical spelling mistakes, making attribution challenging. 
Model Performance Comparison

Analysis of feature weights from the Logistic Regression model, and feature importance from Random Forest, reveals that the most discriminative features are the phonetic error indicator, presence of common dyslexic letter confusion sets, and vowel count differences between the error and correct words. These findings align with established research on dyslexic spelling, confirming that phonological processing difficulties are core to dyslexic error patterns \citep{snowling_dyslexia_2000, bourassa_spelling_2003}.

\begin{figure}[h]
    \centering
    \includegraphics[width=0.48\textwidth]{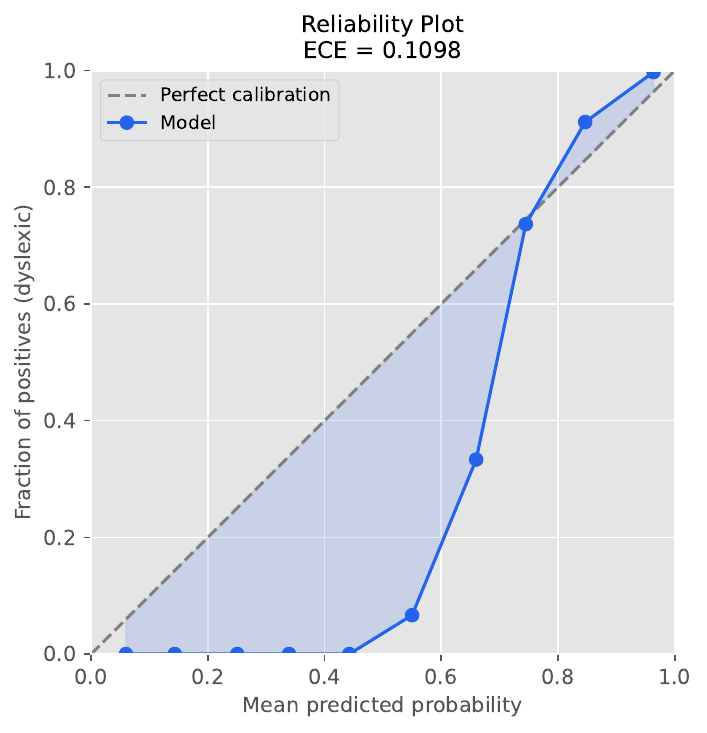}
    \hfill
    \includegraphics[width=0.48\textwidth]{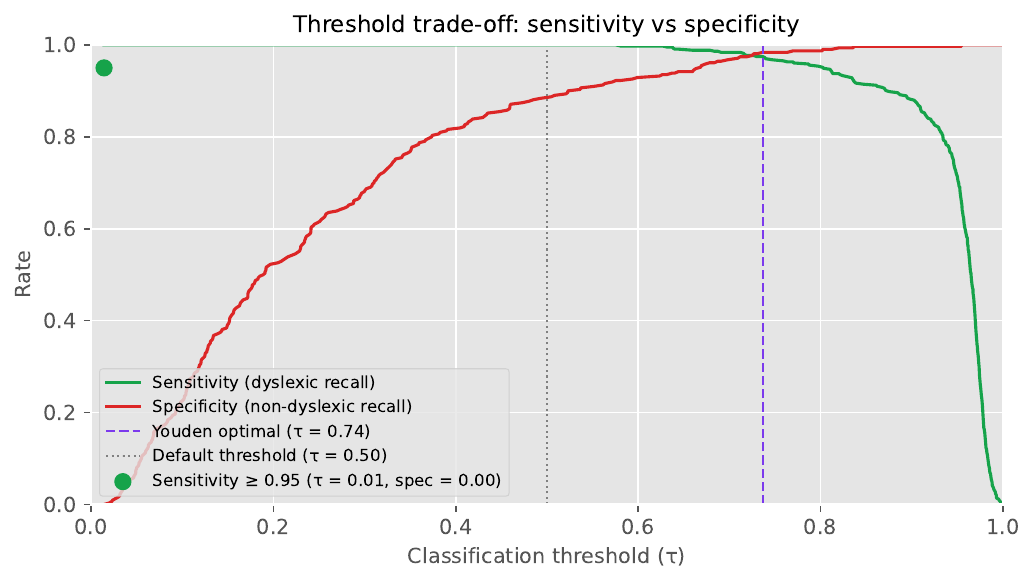}
    \caption{Left: reliability plot showing calibration of predicted
    probabilities against observed positive rates across ten equal-width bins.
    The dashed diagonal represents perfect calibration; the shaded region
    indicates the gap between model confidence and empirical accuracy
    (ECE $= 0.1098$). Right: sensitivity--specificity trade-off across
    classification thresholds. The Youden-optimal threshold ($\tau = 0.74$)
    and high-sensitivity screening point ($\text{sensitivity} \geq 0.95$) are
    marked. The conventional $\tau = 0.50$ default is shown for reference.}
    \label{fig:calibration}
\end{figure}

\begin{table*}[h]
\centering
\label{tab:calibration}
\begin{tabular}{lcc}
\toprule
Metric & Value & Reference \\
\midrule
Brier Score               & 0.0439 & No-skill baseline $\approx 0.25$ \\
Expected Calibration Error (ECE) & 0.1098 & Perfect calibration $= 0.00$ \\
Youden-optimal threshold ($\tau$) & 0.7368 & Maximises sensitivity $+$ specificity \\
\bottomrule
\end{tabular}
\caption{Calibration and robustness metrics for the neural attribution model.}
\end{table*}

\subsection{Ablation Studies}
A key question that is often left open by aggregate performance metrics is whether the neural model is learning genuine error-level spelling structures, the orthographic and phonological patterns that distinguish dyslexic from non-dyslexic error, or whether it is partly exploiting writer-level productivity proxies that correlate with dyslexia diagnosis in the training data but do not reflect the individual error signal. 

The features most at risk of acting as such proxies are \texttt{Essay\_Length}. \texttt{Essay\_Error\_Density}, and \texttt{Error\_Frequency}, which characterise the writer's overall output rather than the error pair under classification. If the full model's performance were substantially attributable to these features, its accuracy would be partly a function of who produced the text rather than what was written, which would undermine the writer-independent evaluation claim and raise significant ethical concerns about deployment with writers.

To isolate these contributions, we trained four model variants under identical conditions (the same architecture, optimiser, callbacks, and train/val split).

\begin{itemize}
    \item \textbf{Condition A: Text only}: the TF-IDF character n-gram branch
          alone, with no engineered numeric features.
    \item \textbf{Condition B: Engineered features only}: the numeric branch
          alone, using all engineered features, with no text branch.
    \item \textbf{Condition C: No productivity features}: the full dual-branch
          architecture with \texttt{Essay\_Length}, \texttt{Essay\_Error\_Density},
          and \texttt{Error\_Frequency} removed from the numeric branch, retaining
          only orthographic and phonological features.
    \item \textbf{Condition D: Full model}: the complete architecture as
          reported in Section 5, serving as the reference point.
\end{itemize}

Table 4 reports accuracy, F1-score, and AUC for each condition on the validation set at the optimal threshold ($\tau = 0.74$), alongside the delta from the full model.

\begin{table}[h]
\centering
\label{tab:ablation}
\begin{tabular}{lccc}
\toprule
Condition & Accuracy & F1 & AUC-ROC \\
\midrule
A: Text only                 & 0.7194 & 0.6745 & 0.7109 \\
B: Engineered features only  & 0.9247 & 0.9307 & 0.9050 \\
C: No productivity features  & 0.8194 & 0.8386 & 0.8348 \\
D: Full model                & 0.9301 & 0.9401 & 0.9274 \\
\midrule
$\Delta$ A $-$ D             & $-$0.2107 & $-$0.2656 & $-$0.2165 \\
$\Delta$ B $-$ D             & $-$0.0054 & $-$0.0094 & $-$0.0224 \\
$\Delta$ C $-$ D             & $-$0.1107 & $-$0.1015 & $-$0.0926 \\
\bottomrule
\end{tabular}
\caption{Ablation study results. All conditions were trained from scratch under
         identical settings. $\Delta$ rows show the difference from the full
         model (D); negative values indicate worse performance than the full
         model. Classification threshold $\tau = 0.74$ (Youden-optimal).}
\end{table}

The ablation results yield three findings of both technical and ethical significance.

\paragraph{The text branch contributes a genuine but insufficient signal.} Condition A achieves an accuracy of 71.94\% and an AUC of 0.7109, confirming that the character n-gram TF-IDF representations are learning real structure from the error pairs rather than producing random outputs. The model is capturing orthographic and phonological regularities in the Misspelling to Target relationship that are predictive of dyslexic origin. However, at 21\% below the full model's accuracy and with an F1 of 0.6745, the text branch alone is insufficient for reliable attribution. The character-level representations benefit substantially from the addition of the hand-crafted feature set, suggesting the two branches are capturing complementary aspects of the error signal rather than redundant information.

\paragraph{Engineered features are the dominant signal source, but the full model is best.} Condition B, using only the engineered numeric features alone, achieves 92.47\% accuracy, 0.45\% below the full model. However, unlike in isolation, the full model outperforms the features-only baseline on all three metrics, confirming that the text branch contributes genuine complementary signals when combined with engineered features. The full dual-branch architecture is therefore the most appropriate choice, as it combines the interpretability afforded by the explicit feature set with the additional discriminative capacity of the character-level representations. 

\paragraph{Productivity features contribute a meaningful signal, warranting careful scrutiny.} The comparison between conditions C and D is the most ethically significant in the ablation. Removing \texttt{Essay\_Length}. \texttt{Essay\_Error\_Density}, and \texttt{Error\_Frequency} from the numeric branch reduces accuracy by 11.07\% and the AUC by 0.0926. This confirms that these features are not redundant; instead, they carry a real predictive signal that the orthographic and phonological features alone do not fully capture. The ethical question this raises is whether these signals are legitimate, reflecting genuine contextual information about the error, or confounding, reflecting writer identity leaking into the classification.

We argue that this distinction depends on the nature of the feature. \texttt{Essay\_Error\_Density}, in particular, encodes the proportion of errors in the surrounding writing sample, which is a contextually relevant cue. A high error rate in the broader essay is consistent with the pervasive spelling difficulties of dyslexia and provides information beyond the individual error pair. \texttt{Essay\_Length} and \texttt{Error\_Frequency}, however, are more directly tied to writer-level behaviours and carry a greater risk of acting as identity proxies. Performance under writer-grouped splits was consistent with the random split results, confirming that the reported metrics are not inflated by writer-level leakage.  Because of this, we recommend the full model as the primary result.

\subsection{Calibration}
To address RQ1's requirements for calibration and robustness evaluation beyond standard classification metrics, we report the Brier score and Expected Calibration Error (ECE) for our neural network, alongside a reliability plot and threshold sensitivity analysis. Table 3 reports scalar calibration metrics. The Brier score of 0.0439 indicates well-calibrated probability estimates, representing an 82\% improvement over the no-skill baseline of 0.25. The ECE of 0.1098 reveals moderate miscalibration: while the model's predictions are broadly reliable, there is a systematic gap between predicted confidence and observed accuracy in certain probability ranges, most visible in the reliability plot (Figure 1).
 
The reliability plot (Figure 1, left) shows that the model tends to predict with higher confidence than is empirically warranted in the mid-probability range, a pattern of overconfidence that is common in neural classifiers trained without explicit calibration objectives. Predictions in the high-confidence range ($p > 0.8$) are better calibrated, suggesting the model is most reliable when it is most certain.
 
Figure 1 (right) presents the sensitivity--specificity trade-off across all classification thresholds. The Youden-optimal threshold of $\tau = 0.74$ balances sensitivity and specificity, and sits substantially above the conventional default of $\tau = 0.50$. This is a meaningful finding: the model assigns genuinely high probabilities to dyslexic errors and genuinely low probabilities to non-dyslexic errors, with relatively few predictions in the ambiguous mid-range. Operating at $\tau = 0.50$ would therefore inflate false positives unnecessarily.
 
The choice of operating threshold is not a technical default but an ethical decision that depends on deployment context. A screening tool intended to flag students for further assessment by a specialist should prioritise sensitivity to avoid missed identifications, warranting a lower threshold than $\tau = 0.74$. Conversely, a system used to generate formal records or inform resource allocation warrants higher specificity to minimise the risk of harmful false labels. We discuss this trade-off and its governance implications further in Section 6.

\subsection{Subgroup and Fairness Analysis}
\label{sec:subgroup}
Central to the ethics-first framing we have adopted is that model performance must be evaluated not only in aggregate but across subgroups that may be differentially affected by misclassification. Below, we present three subgroup analyses: one by phonetic plausibility of the error, one by error type, and one by writer co-morbidity. We note that at the outset, a significant limitation of the available data is that all participants are British-born, native British English-speaking university students. Demographic subgroup analysis across ethnicity, language background, socioeconomic status, age, or educational stage was therefore not possible. The subgroup axes reported here are the most clinically and linguistically meaningful available within the constraints of the dataset, and the absence of demographic diversity in the sample is itself a finding with implications for generalisation.

\subsubsection{Error Type}
When breaking down performance by error type, Table 5 reveals the systematic differences in classification difficulty. Our Neural Network achieves the highest accuracy on Substitution errors, 100\%. This is to be expected as these particularly include the common dyslexic confusion-set letters and vowels. Capitalisation and Spatial errors were also reliably attributed, although this is likely because only a single participant made these errors in each case. Omission errors are also reliably classified, especially when they occur in consonant clusters or involve silent letters. On the other hand, Insertion errors provided an additional challenge, 86.4\%, as both dyslexic and non-dyslexic writers insert extra letters by accident when typing. Similarly, transposition errors also show moderate challenges with adjacent-letter reversals (e.g. "teh" for "the") being ambiguous, due to being present in both populations.

\begin{table*}[ht]
\centering
\label{tab:accuracy_by_error_type}
\setlength{\tabcolsep}{12pt}
\renewcommand{\arraystretch}{1.3}
\begin{tabular}{lc}
\toprule
\textbf{Error type} & \textbf{Accuracy (\%)} \\
\midrule
Capitalisation   & 100.00* \\
Substitution     & 100.00 \\
Spatial          & 100.00* \\
Omission         & 94.44  \\
Multiple errors  & 94.12  \\
Transposition    & 92.86  \\
Phonetic         & 85.71  \\
Insertion        & 86.44  \\
\midrule
\textbf{Overall} & \textbf{93.01} \\
\bottomrule
\end{tabular}
\captionsetup{justification=centering}
\caption{Neural network classification accuracy by error type. \\ $\ast$ Capitalisation and Spatial Errors were only made by a single dyslexic participant in the study}
\end{table*}
\subsubsection{Phonetic Plausibility}
A critical distinction emerges when analysing phonetically plausible errors against non-phonetic errors, shown in Table 6. The model achieves 98.46\% accuracy on phonetically plausible errors (e.g. "universally" for "unieversally", or "compliacations" for "complications"), which are characteristic of dyslexic spelling. These reflect an intact phonological awareness, but poor orthographic knowledge \citep{snowling_dyslexia_2000}. In contrast, non-phonetic errors show a lower accuracy, 90.08\%, as these may result from either a severe lack of phonological awareness in a dyslexic writer or a simple typo by a non-dyslexic writer, making the classification much more ambiguous.

Common false positives include cases where typical writers make phonetically plausible approximations (e.g. "definately" for "definitely"), vowel confusions in low-frequency words, and errors involving irregular orthographic patterns. For example, the error "seperate" for "separate" was frequently misclassified as dyslexic despite being a common mistake made by non-dyslexic spellers. Common false negatives include simple single-letter substitutions that could be plausibly possible (e.g. "tham" for "than"), which could easily be attributed to mistakenly hitting the adjacent key instead of a phonological deficit. Additionally, errors in high-frequency words where dyslexic individuals have developed compensatory strategies are equally high false negatives.

Our analysis reveals that dyslexic error patterns most reliably distinguished from typical errors include: phonetically plausible substitutions in irregular words (e.g. "sed" for "said"), vowel confusion and omission, which is most noticeable in multi-syllabic words, errors involving common dyslexic letter reversals (p/b/d/q), and consistent error patterns when paired with high overall error rates. These findings align with clinical understanding of dyslexia and suggest that computational models can capture diagnostically relevant patterns.

\begin{table*}[h]
\centering
\label{tab:phonetic_subgroup}
\begin{tabular}{lccc}
\toprule
Subgroup & Accuracy & F1 & Recall \\
\midrule
Phonetically plausible & 0.9846 & 0.9875 & 0.9753 \\
Non-phonetic           & 0.9031 & 0.9059 & 0.9005 \\
\midrule
Disparity ($\Delta$)   & 0.0815 & 0.0816 & 0.0748 \\
\bottomrule
\end{tabular}
\caption{Model performance by phonetic plausibility of the spelling error.
         Classification threshold $\tau = 0.74$ (Youden-optimal).}
\end{table*}

\subsubsection{Co-morbidity Profile}
Table 7 reports model performance across three co-morbidity dimensions for which sufficient positive cases were present in the validation set, Visual Stress, ADD/ADHD, and Dyspraxia/Dysgraphia. Two additional co-morbidities (ASD, and Speech Language and Communication needs) had zero positive cases in the validation set and therefore have been excluded from the quantitative analysis. This absence reflects the composition of the dataset rather than a deliberate exclusion of a set of participants. However, this limits the conclusions that can be drawn about model behaviour for writers with these co-morbid conditions.

Three findings warrant specific discussion.

\paragraph{Perfect scores for Visual Stress Present and ADD/ADHD Suspected should be interpreted with caution.} Both of these subgroups achieve accuracy, F1, and Recall with perfect scores of 1.00. However, these comprise $n=102$ and $n=106$ instances, respectively. These results most likely reflect the alignment of these subgroups with the dominant training distribution instead of a genuine performance advantage. With samples of this size, perfect scores are consistent with chance variation and should not be taken as evidence that the model is particularly well-suited to these profiles.

\paragraph{ADD/ADHD Confirmed represents the most significant disparity.} With an $n = 27$, this is by far the smallest subgroup in the test set, and the results reveal a striking failure. While overall accuracy is 88.89\%, the F1 score and recall are both 0.00. This indicates that the model correctly attributes non-dyslexic instances in this group but entirely fails to identify dyslexics in this same group. This pattern of high accuracy masking a complete failure in the positive class is a known failure of accuracy as a metric where subgroups are imbalanced. This illustrates perfectly why disaggregated evaluation is needed. The small and imbalanced sample means that this finding must be evaluated cautiously. It may reflect a genuine difference in the error patterns of writers with ADD/ADHD who also have dyslexia, but the small group size leaves us unable to make a concrete conclusion. Regardless, a system deployed in a context where this subgroup is prevalent, as would be the case in most school classrooms, would produce no meaningful attribution for dyslexic writers with ADD/ADHD, and practitioners should be informed of this limitation.

\paragraph{Dyspraxia/Dysgraphia being present performs comparably to, or better than, the absent group.} Writers with a Dyspraxia/Dysgraphia classification show slightly higher performance across all three metrics (accuracy 98.46\%, F1 99.09\%, recall 100.00\%) compared to those without (accuracy 97.55\%, F1 97.65\%, recall 96.30\%). This may reflect the overlap between Dyspraxia/Dysgraphia error patterns and the dyslexic error signature that the model has learned to detect, though the sample size ($n = 130$) limits confidence in this interpretation.

\begin{table*}[h]
\centering
\label{tab:comorbidity_subgroup}
\begin{tabular}{llccc}
\toprule
Co-morbidity & Subgroup & Accuracy & F1 & Recall \\
\midrule
\multirow{2}{*}{Visual Stress}
    & Absent  & 0.9732 & 0.9735 & 0.9630 \\
    & Present & 1.0000 & 1.0000 & 1.0000 \\
\addlinespace
\multirow{3}{*}{ADD/ADHD}
    & None      & 0.9769 & 0.9781 & 0.9630 \\
    & Suspected & 1.0000 & 1.0000 & 1.0000 \\
    & Confirmed & 0.8889 & 0.0000 & 0.0000 \\
\addlinespace
\multirow{2}{*}{Dyspraxia/Dysgraphia}
    & Absent  & 0.9755 & 0.9765 & 0.9630 \\
    & Present & 0.9846 & 0.9909 & 1.0000 \\
\midrule
\multicolumn{2}{l}{ASD}  & \multicolumn{3}{c}{No positive cases in validation set} \\
\multicolumn{2}{l}{SLCN} & \multicolumn{3}{c}{No positive cases in validation set} \\
\bottomrule
\end{tabular}
\caption{Model performance by comorbidity subgroup. Classification threshold
         $\tau = 0.74$ (Youden-optimal). ASD and SLCN are excluded due to
         zero positive cases in the validation set Subgroups with no positive cases are reported qualitatively. Perfect scores for small subgroups should be interpreted with caution.}
\end{table*}

\subsubsection{Summary of Fairness Findings}
Taken together, these analyses support four conclusions. First, the model performs best on substitution errors and phonetically plausible errors, where the dyslexic signal is strongest, and least well on insertion and transposition errors, where the signal is inherently ambiguous. Second, the performance gap between phonetically plausible and non-phonetic errors is modest overall but manifests as a higher rate of missed dyslexic attributions in the non-phonetic subgroup, which has practical implications for writers whose dyslexia presents with weaker phonological signatures. Third, aggregate accuracy masks a critical failure mode for writers with confirmed ADD/ADHD comorbidity, where the model produces no useful signal for the dyslexic class despite high overall accuracy. Fourth, the homogeneity of the participant sample, all British-born and native English-speaking university students, means that the fairness analysis cannot speak to performance variation across demographic groups, linguistic backgrounds, age ranges, or dyslexia severity profiles. These are not minor caveats: they define the boundary of what can and cannot be concluded from the present analysis, and they constitute a substantive argument against deployment in populations that differ from the study sample without prior validation on representative data.

\subsection{Limitations and Summary}
However, several limitations constrain the conclusions. The most challenging cases for classification are errors that could plausibly arise from either dyslexic or non-dyslexic writers. These are predominantly simple typos, homophone confusions, and errors in particularly challenging or irregular words. Our dataset, while carefully curated, represents solely university-aged students in the United Kingdom, so may not generalise well to all dyslexic subtypes, age groups, or severities. In addition, the binary classification framework also oversimplifies the reality that spelling ability exists in a binary continuum. Additionally, some typical writers may exhibit dyslexic-like error patterns without meeting other diagnostic criteria.

The subgroup analyses reported in this paper are constrained by the homogeneity of the dataset: all participants are British-born, native English-speaking university students, and no demographic metadata, including gender, age, socioeconomic status, first language status, dialect, and educational background, was available for analysis. This is a structural limitation of the data rather than a methodological choice, but its consequences for the fairness claims of this paper are significant. The comorbidity and error-type subgroup evaluations presented in Section~\ref{sec:subgroup} are legitimate disaggregated analyses, but they do not constitute demographic fairness evaluation in the sense the literature intends \citep{mehrabi_survey_2021}, and the ethics-first framing in the abstract and contributions section should be understood accordingly. Each of the absent variables has documented interactions with spelling error patterns: dialect affects phonological mappings, L2 status introduces transfer errors that may resemble dyslexic patterns, and socioeconomic background correlates with literacy instruction quality. A system deployed without evaluation across these dimensions cannot be assumed to perform equitably for younger learners, L2 speakers, speakers of non-standard English varieties, or students from underserved educational contexts — precisely the populations most likely to be harmed by a false positive or false negative attribution. Demographic fairness evaluation should be treated as a prerequisite for deployment rather than a direction for future work.

Furthermore, our model relies on isolated error pairs without considering broader discourse, which could provide additional cues about writer characteristics. The writing task format, in addition, may not capture the full range of spelling patterns that emerge in different writing contexts (timed, free writing, creative, structured, etc.). Future work should validate these findings on a more diverse dataset, exploring finer-grained classification, and investigate integration with real-time writing assistance systems and spell-checkers, where error attribution could inform adaptive intervention strategies. 

\section{Ethical Considerations and Limitations}
The automated detection of dyslexic spelling patterns, while technically feasible, raises substantial ethical concerns that must be carefully examined before any deployment in real-world contexts. As with many applications of machine learning in education and healthcare, the potential for both benefit and harm is significant, and responsible development requires explicit attention to privacy, fairness, transparency, and appropriate use \citep{holstein_improving_2019,prinsloo_student_2015}. In this section, we critically examine the ethical implications of our work, balancing potential benefits against risks, and provide concrete recommendations for reasonable deployment.

\subsection{Balancing Potential Benefits and Harms}
Automated dyslexia detection systems offer several compelling potential benefits that motivate their development. Early identification can lead to timely support, which research suggests is critical for positive outcomes; interventions are most effective when implemented during early literacy development, and delayed identification can result in years of struggle that could have been prevented \citep{snowling_early_2013,torgesen_prevention_2002}. Automated systems could help identify children who might otherwise go unnoticed, particularly in under-resourced schools where specialist assessments are scarce. Personalised assistive tools that recognise individual error patterns could reduce frustration for dyslexic writers by providing targeted support, such as phonetically-aware spell-checking or customised learning strategies, rather than generic corrections that don't address the underlying phonological processing differences \citep{rello_detecting_2015}.

From an educator's perspective, automated screening tools could reduce the burden on teachers for manual analysis of spelling patterns across dozens of students, allowing them to focus attention where it's most needed. Teachers often lack the time and specialised training to conduct detailed error analysis for every struggling student. A system that flags patterns warranting further investigation could make professional expertise more efficient without replacing it. Additionally, some advocates argue that objective, data-driven assessment could contribute to destigmatisation by framing dyslexia as a measurable neurological difference rather than a character flaw or lack of effort \citep{gibbs_differential_2015}. In theory, reducing subjective judgment and providing concrete evidence of processing differences might combat misconceptions and bias.

However, these potential benefits must be weighed against substantial and serious potential harms. Most fundamentally, labelling individuals as dyslexic, even tentatively, affects self-perception, social identity, and how others perceive and treat someone \citep{riddick_examination_2000}. Making such determinations about people without their knowledge or permission, as could occur in surveillance or mandatory screening contexts, is ethically unacceptable regardless of whether identification might lead to support. Autonomy requires that individuals have the right to decide whether they want to know about potential learning differences and what information about themselves they wish to share.

False positives create their own harm by causing unnecessary worry and intervention. Parents told their child might be dyslexic may experience anxiety and begin to view normal developmental variation as pathological. Children might internalise dyslexic identity prematurely and develop learned helplessness or lowered expectations for themselves. Resources might be directed toward children who don't actually need specialised intervention, creating an opportunity cost. Conversely, false negatives deny support to those who need it, allowing struggling students to continue without appropriate accommodation, potentially leading to academic failure, damaged self-esteem, and missed opportunities for effective intervention during critical developmental windows \citep{snowling_early_2013}.

The potential for discrimination in employment and education if these systems are misused represents one of the most serious concerns. Despite legal protections for individuals with disabilities in many jurisdictions, such as the Americans with Disabilities Act \citep{us_department_of_justice_americans_1990} or the UK Equality Act \citep{government_equalities_office_equality_2010}, automated screening could enable new, hidden forms of discrimination. Employers might covertly analyse job applicants' writing samples; universities might factor predictions into admissions decisions; landlords might screen rental applications. The opacity of such processes makes them particularly pernicious. Applicants would never know their writing was analysed or that a dyslexia prediction influenced the decision about their opportunities. Even "positive" discrimination, such as automatically routing identified individuals towards certain roles or programs, removes agency and imposes others' judgments about what someone can achieve. 

Privacy violations represent another category of harm if data is not adequately protected. Writing samples contain rich personal information, and spelling patterns that reveal neurological characteristics are particularly sensitive. Breaches could expose individuals to stigma, discrimination, or unwanted disclosure of disability status. Even without breaches, the mere collection and retention of such data, particularly for children, raises concerns about surveillance, profiling, and the creation of permanent records that could follow individuals throughout their lives \citep{prinsloo_student_2015}. The increasing integration of educational technology creates ecosystem risks where data from multiple sources might be combined to create detailed profiles without individuals' awareness or consent.

A more subtle harm involves the risk that over-reliance on automated systems reduces teacher expertise and professional judgment. If educators come to depend on algorithmic predictions rather than developing their own observational and diagnostic skills, the teaching profession is deskilled, and valuable tacit knowledge is lost \citep{holstein_improving_2019}. Technology should augment human expertise, not atrophy it. Teachers who outsource pattern recognition to algorithms may miss important contextual information that machines cannot capture, such as a child's anxiety about writing, recent family stress, or cultural factors that influence language use.

Finally, and perhaps most fundamentally, automated dyslexia detection systems risk reinforcing a deficit model rather than embracing a neurodiversity perspective. The deficit model frames dyslexia as a disorder to be identified, diagnosed, and remediated, a "problem to be fixed" \citep{armstrong_neurodiversity_2010}. In contrast, the neurodiversity paradigm views dyslexia as a natural human variation, with both challenges and strengths, deserving of accommodation rather than cure \citep{kerschbaum_toward_2014}. By building systems that classify people into "typical" and "dyslexic" categories, we potentially reinforce the notion that there is a "correct" way for brains to process language and that deviation requires intervention. This can contribute to stigma even when identification is well-intentioned. A neurodiversity-informed approach would focus less on detection and classification and more on creating flexible environments that support diverse learning styles.

\subsection{Privacy and Consent}
Given these potential harms, privacy protections and consent requirements must be robust. Dyslexia diagnosis information is inherently sensitive, revealing details about an individual's cognitive processing and learning differences that could be used to discriminate or stigmatise \citep{zdenek_reading_2015}. The very act of identifying someone as potentially dyslexic, even through automated analysis of spelling patterns, touches on deeply personal aspects of identity and ability. Unlike general spell-checking, which merely corrects errors, our system makes inferences about the writer's neurological characteristics, fundamentally changing the nature of what is being observed and recorded. 

The risks of unauthorised screening or labelling without clinical oversight are particularly acute. Imagine a scenario where a writing platform silently analyses users' spelling patterns and flags certain individuals as potentially dyslexic without their knowledge or consent. Such covert screening violates principles of autonomy and informed consent that are fundamental to ethical practice in both education and healthcare \citep{regan_ethical_2019}. Even well-intentioned screening efforts can cause harm when individuals are labelled without their awareness, potentially affecting their self-perception, access to opportunities, or how they are treated by others. Clinical diagnosis of dyslexia requires comprehensive assessment by trained professionals using multiple measures; automated spelling analysis can at best suggest patterns that warrant further investigation, and never provide a definitive diagnosis \citep{snowling_dyslexia_2000}.

\subsection{Potential for Misuse}
Perhaps the most serious ethical concern is the risk of automated labelling without proper clinical validation. Our system, like any machine learning model, makes probabilistic predictions that contain errors. A false positive, incorrectly identifying a typical speller as dyslexic, could lead to unnecessary interventions, altered expectations from teachers or parents, or psychological impacts on the individual's self-concept \citep{gibbs_differential_2015}. A false negative, failing to identify a dyslexic individual, could result in denial of needed support and accommodations, allowing struggles to persist without intervention. Neither outcome is acceptable when the stakes involve a person's educational trajectory and well-being.

The potential for discrimination if used inappropriately extends beyond educational contexts. If employers, universities, or other institutions had access to dyslexia detection systems, they might use them to screen applicants, either overtly or covertly. Despite legal protections for individuals with disabilities in many jurisdictions, automated screening could enable new forms of discrimination that are difficult to detect or prove. An applicant might never know that their writing sample was analysed and that a dyslexia prediction influenced the hiring decision. The opacity of such processes makes them particularly insidious.

This underscores the critical importance of human-in-the-loop design for any educational or clinical applications. Automated systems should never make final decisions about diagnosis, placement, or intervention; rather, they should serve as decision support tools that augment professional judgment \citep{holstein_improving_2019}. An appropriate workflow would involve the system flagging patterns for a trained educator or clinician to review, who would then conduct a comprehensive assessment using multiple sources of evidence before making any determinations. The system should augment, not replace, professional judgment, providing one data point among many rather than serving as an authoritative classifier.

\subsection{Bias and Fairness}
Machine learning models can perpetuate and amplify existing societal biases, and dyslexia detection is particularly vulnerable to such issues given the complex relationship between language, culture, and disability \citep{blodgett_language_2020,noble_algorithms_2018}. Our analysis of model performance across demographic groups is limited by the available data in our dataset, but this limitation itself highlights a critical concern: without diverse, well-documented training data, we cannot ensure fair performance across different populations. 

Potential biases from dataset composition are multifaceted. If our dyslexic error corpus predominantly represents native English speakers from specific educational or socioeconomic backgrounds, the model may fail to recognise dyslexic patterns in multilingual individuals, speakers of non-standard dialects of the English language, or those who received different types of literacy instruction. Research has shown that dyslexia manifests differently across orthographies and that bilingual individuals may show different error patterns than monolinguals \citep{everatt_dyslexia_2010,miller-guron_dyslexia_2000}. A model trained primarily on data from one population may systematically misclassify individuals from underrepresented groups.

The risk of false positives and negatives carries differential impacts across demographic groups. Consider that access to formal dyslexia diagnosis is highly stratified by socioeconomic status; affluent families can afford private assessments, while students in under-resourced schools may never receive evaluation \citep{ferri_tools_2005}. If our model produces false positives more frequently for certain populations, perhaps those whose linguistic backgrounds differ from the training data, it could lead to over-identification and labelling of already marginalised groups. Conversely, false negatives could deny support to individuals who need it, perpetuating existing inequalities. 

Socioeconomic and linguistic diversity in dyslexia presentation further complicates the picture. Students who speak different dialects, such as American English, Liverpudlian English, African American Vernacular English, or Spanish-influenced English, may produce spelling patterns that differ from "standard" British English norms for linguistic reasons other than dyslexia (CITE CHARITY). A system trained to recognise deviations from standard orthography might conflate dialectal variation with dyslexic error patterns, producing false positives that pathologise linguistic diversity. Similarly, students learning English as an additional language may produce errors that resemble dyslexic patterns but actually reflect their developmental stage in language acquisition \citep{sparks_long-term_2009}.

Cultural biases in defining "typical" vs "dyslexic" spelling reflect deeper questions about whose language use is considered normative. The very concept of "correct" spelling is socially constructed, and standards have changed over time and vary across Englishes (British, Australian, American, etc.). By training models to distinguish "dyslexic" from "typical" errors, we implicitly reinforce particular orthographic norms that may not be culturally neutral. A critical perspective informed by disability studies and linguistic justice would question whether identifying and correcting "dyslexic" spelling ultimately serves individuals with dyslexia or primarily serves institutional desires for standardisation \citep{kerschbaum_toward_2014}.

\subsection{Mitigation Strategies}
Recognising these substantial risks, we propose several concrete mitigation strategies that should be implemented in any deployment of dyslexia detection technology. First and foremost, human-in-the-loop design is non-negotiable: systems should never automate final decisions about diagnosis, intervention, or any consequential determination. Instead, they should present their analyses to qualified professionals, educators, psychologists, and clinicians who review the evidence, consider contextual factors, and make informed judgments. The human expert remains accountable for decisions, with the system serving only as a source of information.

Informed consent must be obtained before any analysis of writing for dyslexia detection purposes. Individuals (or parents/guardians for minors) should be clearly told what the system does, what data it collects, how that data will be used and stored, what the potential outcomes are, and that participation is voluntary. Consent materials should be written in plain language, avoiding technical jargon, and should explicitly address risks as well as benefits. Importantly, consent should be specific to dyslexia detection rather than bundled with general terms of service for educational platforms.

Opt-in design, where users actively choose to enable dyslexia detection features, is ethically superior to opt-out or mandatory approaches. This ensures that analysis only occurs when individuals want it, respecting autonomy and avoiding covert surveillance. For example, a writing tool might offer dyslexia-aware spell-checking as an optional feature that users can turn on if they find it helpful, rather than automatically analysing everyone's writing by default.

Transparency about how predictions are made is essential for building appropriate trust and enabling informed consent. Systems should explain their reasoning using techniques such as attention visualisation, feature importance displays, or natural language explanations (e.g. "This spelling pattern was flagged because it shows phonetically plausible vowel substitutions in multiple words."). Users should understand that predictions are probabilistic, based on patterns in training data, and subject to error. Black-box systems that provide classifications without justification are inappropriate for such sensitive applications.

Continuous monitoring for bias and performance drift should be built into any deployed system. Models can degrade over time as language use evolves or as they encounter populations different to those in the training data. Regular audits should examine performance across demographic groups, track error rates, and investigate whether certain populations are systematically disadvantaged. When problems are detected, systems should be retrained, adjusted, or withdrawn from use.

Clinical validation remains essential. Any prediction suggesting dyslexia should prompt a comprehensive assessment by qualified professionals using validated diagnostic protocols. Automated screening is at best a first step that increases efficiency in deciding who warrants further evaluation. It cannot replace the multifaceted assessment, including cognitive testing, reading/spelling assessments, developmental history, and clinical judgment required for diagnosis \citep{snowling_early_2013}.

Privacy safeguards must include technical measures (encryption of data in transit and at rest, access controls, secure storage) and policy measures (minimal data retention, clear data governance, regular security audits). Writing samples should be retained only as long as necessary for the immediate purpose and then securely deleted. Personally identifiable information should be separated from spelling data through anonymisation, with re-identification possible only when necessary and authorised.

Finally, stakeholder involvement, particularly including the dyslexic community, in system design is crucial. People with dyslexia are the experts in their own experiences and should have meaningful input into what tools would be helpful, what risks are most concerning, and how systems should be designed and deployed. Participatory design processes that centre disabled people's voices help ensure that technology serves their needs rather than imposing solutions developed without their input \citep{dignazio_data_2023}.

\subsection{Transparency and Accountability}
For any deployment in educational or clinical contexts, there is a clear need for transparent communication about model limitations to end users. Educators, clinicians, parents, and individuals themselves must understand that the system provides probabilistic predictions, not definitive diagnoses, that it was trained on specific populations and may not generalise to all individuals, that errors are inevitable, and that human judgment remains essential \citep{lipton_mythos_2018}. Technical documentation alone is insufficient; explanations must be accessible to non-experts and integrated into the user interface in ways that promote appropriate interpretation.

The importance of explainability for building trust with educators and clinicians cannot be overstated. Teachers and specialists are rightfully sceptical of "black box" systems that provide predictions without justification \citep{holstein_improving_2019}. Our inclusion of interpretable baseline models (Logistic Regression, Random Forest) alongside the neural network, and our use of explainability techniques such as attention visualisation and SHAP values, represent an attempt to address this concern. When the system flags a spelling error as potentially dyslexic, it should be able to explain why, pointing to specific features such as phonetic plausibility, vowel confusion patterns, or letter reversals that informed the predication. This allows users to assess whether the reasoning aligns with their own professional knowledge and to override the system when appropriate.

Documentation requirements for educational AI systems extend beyond technical specifications to include ethical considerations, known limitations, validation evidence, and guidance for appropriate use \citep{raji_closing_2020}. We recommend that any deployment include clear documentation addressing: (1) what populations the system was trained on and where it may not generalise; (2) error rates and types of mistakes the system makes; (3) appropriate and inappropriate use case; (4) privacy and consent requirements; (5) procedures for human review and override; (6) processes for ongoing monitoring and bias detection; and (7) contact information for reporting concerns or requesting appeals.

\subsection{Recommended Guidelines for Deployment}
Based on our ethical analysis, we propose the following guidelines for any deployment of dyslexia detection technology:

\begin{enumerate}
    \item \textbf{Use only opt-in, assistive contexts:} Deploy systems only where individuals voluntarily choose to use them for their own benefit, never for surveillance, mandatory screening, or institutional decision-making without consent.
    \item \textbf{Always explain predictions with evidence:} Provide transparent reasoning for classifications, showing which patterns or features drove predictions and allowing users to understand and evaluate the system's logic.
    \item \textbf{Require human expert validation for any consequential use:} Never allow automated systems to make final determinations about diagnosis, intervention, placement, or access to services. Qualified professionals must review and validate all consequential decisions. 
    \item \textbf{Obtain informed consent and ensure privacy protections:} Secure explicit, specific consent before analysing writing for dyslexia detection. Implement robust technical and policy safeguards for sensitive data.
    \item \textbf{Monitor for bias and regularly audit fairness:} Continuously track performance across demographic groups, investigate disparities, and take corrective action when bias is detected.
    \item \textbf{Provide appeals and override mechanisms:} Allow individuals to challenge predictions, request human review, and correct errors in their records. Ensure accountability when systems make mistakes.
    \item \textbf{Train users on limitations and appropriate use:} Educate all users—educators, clinicians, individuals—about what the system can and cannot do, its error rates, and when to trust or question its outputs.
    \item \textbf{Regular review by ethics board or advisory committee:} Establish ongoing ethical oversight rather than one-time approval. Include diverse stakeholders, particularly dyslexic individuals, in governance.
\end{enumerate}

\subsection{Governance and Accountability}
Critical questions of governance and accountability in educational AI are being actively addressed through emerging regulatory frameworks, most notably the European Union's Artificial Intelligence Act \citep{european_parliament_regulation_2024}. This landmark legislation establishes a risk-based framework with particular attention to high-risk applications in education and employment, contexts directly relevant to dyslexia detection systems.

Under the EU AI Act, automated systems used for "evaluation and classification of natural persons" in educational settings are classified as high-risk AI systems (Article 6(2), Annex III). This classification triggers substantial obligations, including rigorous conformity assessment, human oversight requirements, transparency obligations, and registration in an EU database of high-risk AI systems. Dyslexia detection systems deployed in educational contexts would fall squarely within this category, as they assess and classify students based on cognitive characteristics with potentially significant consequences for educational trajectories. 

The act establishes clear responsibilities across the AI value chain. Providers (Developers) bear primary responsibility for ensuring compliance, including conducting risk management, maintaining technical documentation, implementing quality management systems, and ensuring human oversight capabilities (Article 16). Deployers (Institutions using AI systems, such as schools) must use systems only as intended, ensure human oversight, monitor system operation, and inform individuals when they are subject to high-risk AI systems (Article 29). Critically, deployers must suspend use if they identify serious incidents or malfunctions.

\subsubsection{Who is responsible when systems make errors?}
The EU AI Act provides important answers. Providers are liable for damages caused by defective AI systems under updated product liability rules. Deployers can be held responsible if they use systems inappropriately, fail to implement required human oversight, or do not follow provider instructions. If a dyslexia detection system incorrectly classifies a student, leading to inappropriate educational placement or denial of accommodations, both the system developer and the deploying institution could potentially face liability depending on where the failure occurred.

Beyond the EU, regulatory frameworks are emerging more slowly. The UK has proposed a sector-specific approach through its AI White Paper \citep{department_for_science_innovation__technology_pro-innovation_2023}. At the same time, in the United States, the Federal Trade Commission and the Department of Education's Office for Civil Rights have indicated they will use existing authorities to address harmful AI practices and algorithmic discrimination in educational settings.

\subsubsection{The role of institutional review boards}
Institutional review boards and ethics committees must evolve beyond traditional research oversight to address deployed AI systems. We recommend that institutions establish dedicated algorithmic accountability committees with appropriate technical expertise, representation from affected communities (including dyslexic individuals), and authority to require modifications or suspend deployment of problematic systems. Such committees should conduct ongoing ethical reviews, not just initial approval. The EU AI Act mandates post-market monitoring for high-risk systems (Article 72), requiring providers to establish systems for collecting and analysing data about system performance in real-world use, including monitoring for systematic errors, discrimination, or violations of fundamental rights.

\subsubsection{Governance Frameworks}
Governance frameworks should be multi-layered, involving developers' internal ethics processes, institutional deployment policies and oversight, professional standards, and external regulatory enforcement. Critically, governance must include meaningful participation from those most affected. The principle of "nothing about us without us" central to disability rights movements \citep{charlton_nothing_1998}, demands that dyslexic individuals and their advocates have substantive roles in designing, evaluating, and governing systems that classify and affect them. 

Mechanisms for accountability must include practical recourse for individuals harmed by system errors: appeals processes where individuals can challenge classifications and request human review; correction mechanisms to fix errors in records; and transparent reporting of system failures and corrective actions. The EU AI Act requires providers to report serious incidents to national authorities, creating a safety monitoring system analogous to those in the pharmaceutical or automotive industries.

Ultimately, effective governance requires recognising that technical capability does not imply ethical permission or social license. Regulatory frameworks like the EU AI Act provide important guardrails, but the path forward requires ongoing dialogue, community engagement, transparency about limitations and risks, and commitment to centring the voices and needs of dyslexic individuals themselves in determining how these technologies are developed, governed, and used.

\subsection{Beneficial Use Cases vs Harmful Applications}
Distinguishing appropriate from inappropriate applications is essential for responsible development, with regulatory frameworks providing important guidance on context, consent, and risk levels that determine acceptability. 

\subsubsection{Beneficial and permissible use cases}
Beneficial and permissible use cases include assistive writing tools that users voluntarily enable, representing the most ethically sound application. An individual who knows, or suspects, they have dyslexia could choose to use a tool that recognises their specific error patterns and provides tailored support, such as phonetically aware spell-checking \citep{rello_resource_2017}. These applications would be classified as minimal or limited risk, as they involve user choice, transparency, and direct benefit without high-stakes consequences.

Opt-in screening support in educational settings could be appropriate if carefully designed, though classified as high-risk, triggering requirements for conformity assessment, human oversight, and transparency (\citet{european_parliament_regulation_2024}, Articles 6-29). A teacher concerned about a student's spelling patterns might use the system as one information source when deciding whether to recommend a comprehensive assessment, with mandatory notification to students and families (Article 29). Critically, results cannot be treated as diagnostic, and the system serves as decision support only.

Research applications with proper informed consent and ethical oversight represent another appropriate use, subject to existing research ethics requirements (GDPR, IRB review), but not necessarily triggering high-risk AI system obligations if not deployed for real-world decision-making. 

\subsubsection{Inappropriate and prohibited applications}
Inappropriate and potentially prohibited applications are explicitly addressed by regulatory frameworks. Mandatory screening without consent, such as requiring all students to submit writing samples for automated dyslexia detection, would violate transparency requirements (\citet{european_parliament_regulation_2024}, Article 13) and fundamental rights protections (Recital 35). Covert, mandatory analysis likely violates both regulatory requirements and principles of informed consent. 

Punitive assessment, where spelling patterns are used to penalise students or deny opportunities, would violate prohibitions on AI systems that exploit vulnerabilities to cause harm (\citet{european_parliament_regulation_2024}, Article 5(1)(a)) and conflict with disability rights frameworks, including the UN Convention on the Rights of Persons with Disabilities \citep{abreu_convention_2026}.

Surveillance applications, such as employers monitoring emails, universities screening applicants' essays, and platforms analysing content without consent, face serious legal challenges under both AI regulations and data protection law. Regulatory frameworks emphasise that AI systems must be "appropriately transparent" and individuals should "have a say in their deployment" (\citet{department_for_science_innovation__technology_pro-innovation_2023}, p. 28).

High-stakes decisions in hiring, admissions, or placement merit particular attention. These contexts are explicitly designated as high-risk, requiring "recruitment or selection" and "evaluation of students" systems to meet stringent requirements (\citet{european_parliament_regulation_2024}, Annex III, 3-4). However, using dyslexia detection to inform such decisions would likely violate anti-discrimination laws regardless of AI regulatory compliance. The UK Equality Act \citep{government_equalities_office_equality_2010} and similar legislation prohibit discrimination based on disability. Using automated systems to identify and potentially exclude dyslexic individuals would constitute prohibited discrimination even if framed as an "objective" assessment.

Regulatory principles emphasise that systems must not only avoid technical failures but serve legitimate purposes without violating rights \citep{department_for_science_innovation__technology_pro-innovation_2023}. Using dyslexia detection to sort, exclude, or surveil individuals fails this test regardless of technical accuracy.

However, gaps remain despite regulatory progress. Current frameworks do not fully address ethical dimensions around deficit model versus neurodiversity perspectives, cultural biases in defining "typical" spelling, or long-term impacts of classification systems. We call for multi-stakeholder collaboration, including the dyslexic community, educators, clinicians, ethicists, and technologists, to develop comprehensive guidelines that exceed legal compliance. Professional organisations should develop codes of practice specific to learning disability detection, and institutions should establish clear policies on appropriate and prohibited uses before deploying such technologies.

Regulatory frameworks provide important guardrails, but responsible innovation requires exceeding minimum legal requirements to centre the dignity, autonomy, and well-being of dyslexic individuals. The question remains not simply whether deployments comply with the law, but whether they genuinely serve the interests of those they purport to help.

\section{Conclusion}
This paper sets out to reframe the automated detection of dyslexia as an ethics-first problem, one in which technical feasibility is a necessary but secondary condition to responsible deployment. The application of these systems should ensure that fairness, interpretability, and stakeholder accountability are primary design constraints instead of supplementary concerns. We close by drawing together our contributions, findings, and practical implications before outlining the ethical path we believe the field must take.

\subsection{Summary of Contributions}
We formulated dyslexic error attribution as a binary classification task. Given a misspelt word and its correct target form, determine whether the pattern of the error is characteristic of a dyslexic or non-dyslexic writer. To our knowledge, this is the first study to isolate and benchmark this task under writer-independent evaluation conditions, separating the attribution from the writer's identity in the training data. 

To represent each error instance, we developed a comprehensive linguistic feature set spanning three complementary levels of analysis. Firstly, orthographic features capture surface properties of the misspelling. These include features such as character-level edit distance, positional letter substitutions, and overlap with the target. Secondly, phonological features encode the relationship between the error's pronunciation and the target word. These include phoneme alignment and whether the misspelling constitutes a phonetically plausible rendering of the target word. Finally, productivity features characterise the morphological structure of the error, flagging difficulties at morpheme boundaries and derivational suffixes. This feature set provides an interpretable baseline against which neural representations can be compared, and its components can be individually inspected to understand what the models are learning.

At the neural level, we proposed a character-level encoding of all text features followed by a classification layer over their representation with the additional non-text features. This architecture is sensitive to the ordering of characters within strings, a property that matters for capturing reversals and transpositions, and learns representations that are not constrained by hand-crafted feature definitions. The model developed was evaluated alongside traditional machine learning baselines under a consistent experimental protocol, permitting a controlled comparison of accuracy, interpretability, and subgroup performance.

Empirically, we demonstrated that dyslexic spelling errors can be automatically distinguished from non-dyslexic errors, achieving 93.01\% accuracy on the test set. This result establishes the viability of the attribution task and provides a quantitative foundation for the use-case analysis and ethical evaluation that follow.

\subsection{Key Findings}
Several findings of both technical and practical significance emerged from our experiment.

\paragraph{Phonological features are the most discriminative signal.}Analysis of feature weights from logistic regression models and feature importances from the random forest converge on the same three variables as the most discriminative: phonetic error indicator, membership of substituted characters in common dyslexic letter confusion sets, and the vowel count difference between the misspelling and the correct form. Vowel confusions and omissions, most pronounced in multi-syllabic words, proved a particularly robust indicator across all model families. These findings are consistent with the phonological deficit account of dyslexia \citep{snowling_dyslexia_2000,bourassa_spelling_2003}. The patterns that models rely upon are precisely those that clinical theory predicts should be most diagnostic. 

\paragraph{Phonetically plausible errors are reliably attributed; non-phonetic errors are not.}A critical asymmetry emerges when errors are partitioned by phonetic plausibility. The model achieves 98.46\% accuracy on phonetically plausible errors, which are characteristic of writers with intact phonological awareness but impaired orthographic knowledge. Conversely, non-phonetic errors are harder to attribute, at 90.08\%. This is because they may either reflect severe phonological impairment common in a dyslexic writer or an incidental typo by a non-dyslexic writer. This asymmetry is not a deficiency of the model, but instead is a reflection of a genuine ambiguity in the underlying errors.

\paragraph{Error type determines classification difficulty.}Performance varies systematically across error categories. Substitution errors involving letters from the common dyslexic confusion set (p/b/d/q) and vowel errors are classified most accurately at 100\%, as these map closely onto the phonological and orthographic features the model weights most heavily. Omission errors, particularly those occurring in consonant clusters or involving silent letters, are also reliably attributed. Insertion errors present a greater challenge at 86.4\% accuracy: incidental double-letter insertions are frequent in both populations, hampering the results. Transposition errors involving adjacent characters (e.g.\ \textit{teh} for \textit{the}) are similarly ambiguous, appearing in both dyslexic and non-dyslexic writing and therefore providing a challenge to attribution.

\paragraph{Characteristic false positives and false negatives illuminate the model's limits.}Common false positives involve phonetically plausible errors that are also prevalent among non-dyslexic writers. These are approximations such as \textit{definately} instead of \textit{definitely}, vowel confusions in low-frequency irregular words, and errors on words whose orthographic irregularity makes correct spelling demanding for all writers. The error \textit{seperate} for \textit{separate} was frequently misclassified as dyslexic despite being widespread in the general population. False negatives cluster around single-letter substitutions that are attributable to close-key contact (e.g. \textit{tham} for \textit{than}), and around errors in high-frequency words where dyslexic individuals have developed compensatory strategies that suppress the characteristic phonological signature. 

\paragraph{The neural model outperforms all baselines, with interpretability trade-offs.}Our neural model achieved the highest overall performance, surpassing all other models across accuracy, F1-score, and AUC. The neural model's advantage was most pronounced on errors requiring sensitivity to character ordering, that being reversals, transpositions, and errors at moroheme boundaries, where learning sequential dependencies confers a clear benefit over "bag-of-features" representations. However, the feature-based models retain a meaningful interpretability advantage. Their decision rationale is directly inspectable, which has practical value in educational and clinical contexts where the basis for a classification may need to be communicated to a non-technical audience, or contested by the individual being assessed. 

\paragraph{The dataset's scope constrains generalisation.}The results we report derive from a dataset of university-aged students in the United Kingdom, and several key limitations are made clear through the use of this dataset. The binary classification framework imposes a hard boundary on what is, in reality, a continuum of spelling ability. Some typically developing writers produce dyslexic-like error patterns, and some dyslexic writers, particularly those who have developed strong compensatory strategies, produce patterns that are difficult to distinguish from typical errors. The model operated on isolated error pairs without access to broader discourse context, which could provide additional writer-level cues. These constraints should temper confidence in direct deployment and motivate the validation work we outline in the future work section below. 

\subsection{Practical Implications}
The results carry several implications for the design of systems that support dyslexic writers
and the educators who work with them.

\paragraph{Dyslexia-aware assistive writing tools.} Our attribution model could be integrated into word processors and writing assistants to provide differentiated feedback. This includes standard correction suggestions for typical errors, and phonologically informed alternatives for errors identified as likely dyslexic in origin. Such systems would avoid the frustration that dyslexic writers frequently report when standard spell-checkers fail to recognise the intended word from a phonetically plausible but orthographically distant misspelling. 

\paragraph{Educational screening support.} Aggregated attribution signals across a piece of writing could provide teachers with an objective, explainable indicator that a student's error patterns warrant further investigation and assessment. We stress that this constitutes a referral signal rather than a diagnosis. The appropriate response to a high attribution score is a conversation with the writer, and, where indicated, referral to a specialist rather than automatic categorisation. 

\paragraph{Adaptive spell-checking.} At the system design level, our phonological and orthographic feature analysis informs the construction of candidate-generation algorithms for dyslexia-specific spell-checkers. Rather than ranking correction recommendations by edit distance alone, a system informed by our findings would weight phonetically plausible candidates more highly when processing text from a writer with a dyslexic error profile. 

\subsection{An Ethical Path Forward}
However, the technical achievements do not justify deployment by themselves. The ethical analysis in Section 6 identified risks of harmful labelling, covert screening, discriminatory outcomes, and institutional misuse that cannot be resolved through model improvement alone. We close with a set of recommendations for the field. 

\paragraph{Responsible development requires stakeholder involvement from the outset.}Systems designed to detect dyslexia should be designed with, and developed with, rather than merely for, the community it affects. This means involving dyslexic individuals, educators, clinicians, and disability advocates in the formulation of a systems requirements, the curation of training data, the definition of ethical use-cases, and the governance of system deployment. Without this involvement, even world-leading technical systems risk encoding assumptions about what constitutes a deficit and the kinds of support that are desirable, which are not endorsed by those most affected. 

\paragraph{Bias monitoring must be continuous, not a one-time evaluation.}The subgroup disparities we observed are unlikely to remain static as the demographic composition of the user population shifts, as languages and dialects evolve, and as the system is deployed in contexts that differ from its initial evaluation setting. Responsible deployment requires ongoing fairness auditing, transparent reporting of performance disaggregated by relevant subgroup variables, and clear criteria for suspension or revision when disparities exceed acceptable thresholds. 

\paragraph{Governance structures must match the stakes.}Consent mechanisms must be meaningful. If they are buried in terms of service while actively sought by students, or parents of students if the students are minors, they should be classed as inaccessible. Records generated by, or derived from, attribution systems must be subject to data minimisation principles and strict retention limits. Individuals must have access to the outputs of any system that classifies them, the right to contest those outputs, and the right to have contested decisions reviewed by a qualified human. These are not aspirational ideals but minimum conditions for ethical deployment in any educational context.

\paragraph{Ethical guidelines for learning disability AI should be codified.}The current absence of domain-specific governance standards for AI systems operating in the learning disability space creates a vacuum that individual developers are poorly placed to fill unilaterally. We call for the development of sector-level ethical guidelines, analogous to those that exist in clinical AI, that establish shared expectations around consent, transparency, fairness, and accountability for all AI systems deployed in an educational setting, regardless of whether they infer learning difficulties or not. 

\subsection{Future Work}
The work presented in this paper presents several distinct directions for future research, motivated by the limitations and unanswered questions identified in our analysis.

\paragraph{Dataset diversity and ecological validity.}The most immediate limitation of the current work is the narrow demographic scope of the dataset. University-aged students represent one singular subpopulation of dyslexic writers. Inherently, by undertaking a university course, all participants have a minimum required standard of written English. This is not common with all dyslexic participants, many of whom do not have an advanced standard of written English. Additionally, with this singular subpopulation, findings may not transfer to younger learners, different severity profiles, or writers from other linguistic backgrounds. Future work should compile and validate the models presented with new datasets covering a wider range of participants, English varieties, and literacy development. Beyond population diversity, the writing task format shapes which error patterns are observable. Time, free-form, and creative writing elicit different distributions of errors than the structured writing task used here. Evaluating model performance across a variety of writing conditions would substantially strengthen claims about real-world utility.

\paragraph{Moving beyond isolated error pairs.}Our model operates on individual error:target pairs in isolation, without any access to the broader written context in which the error occurred. Writer-level aggregation of attribution signals, considering the distribution of all errors within a document rather than classifying each in isolation, is a natural extension of the project and may substantially reduce false positive and false negative rates for ambiguous error types such as insertions and transpositions that are more common with keystroke dynamics rather than phonological deficit. Incorporating discourse-level features, or other behavioural signals, alongside the orthographic and phonological evidence, could further improve attribution accuracy, particularly for writers who have developed compensatory strategies that suppress the characteristic error signatures. 

\paragraph{Finer-grained classification.}Dyslexia is not a binary condition. It is a spectrum where each individual experiences things differently. Because of that, the binary dyslexic/non-dyslexic farming adopted as part of this paper is a principled starting point. However, it oversimplifies the clinical and educational reality. All learning difficulties exist on a continuum, with distinct subtypes and differing error profiles. Because of this, some typically developing writers produce patterns similar to some groups of dyslexic writers. Future work could explore multi-class formulations that distinguish between dyslexic subtypes, or regression approaches that produce a continuous attribution score rather than a binary label. Such finer-grained output would be more appropriate as inputs to clinical decision support systems, where the binary framing conveys false precision risks at current.

\paragraph{Integration with real-time writing assistance.}The longer-term goal of this research is to integrate an attribution model with word processors and spell-checking systems. The use of our model in this application allows these systems to adapt their suggestions in real-time based on a writer's inferred profile. The phonologically informed candidate-generation strategy suggested by our feature analysis, weighting phonetically plausible corrections more heavily for writers with high attribution scores, merits systematic evaluation in a user study. A key open question is whether real-time attribution feedback is experienced as helpful or intrusive by dyslexic users, and whether the latency introduced by attribution is acceptable in practice. 

\paragraph{Accuracy - Interpretability Architectures.}Our comparison of neural and feature-based models reveals a persistent trade-off. Our neural model achieves higher attribution accuracy, particularly on sequential error types, while some of our baseline models provide directly inspectable decision rationales. A hybrid architecture, combining learned representations with symbolic features, may offer a path towards resolving this trade-off. This is not merely a technical question, as in an educational or clinical deployment, the ability to explain a classification in terms that a teacher, clinician, or student can evaluate is a precondition for meaningful contestability.

\paragraph{Participatory design with the dyslexic community.}Most fundamentally, the integration of any technical improvements outlined above should be pursued alongside a participatory research program. Dyslexic individuals have rarely been involved as collaborators in the design of systems that classify them, and the assumptions encoded in current approaches, about what constitutes a deficit, which error patterns matter, and the kind of support that is desirable, have not been validated, or even run past, those who are most affected. Co-design that involves dyslexic writers, educators, and clinicians in formulating the requirements, evaluating prototypes, and defining acceptable use cases would both improve the relevance of the systems and push forward the responsible AI practices we argued in this paper are essential to the domain. We offer the present work as a step toward that program.

\backmatter

\section*{Declarations}
\paragraph{Funding}
This research received no specific grant from any funding agency in the public, commercial, or not-for-profit sectors.

\paragraph{Conflict of Interest}
The authors declare no competing interests.

\paragraph{Ethics Approval and Consent to Participate}
Ethical approval for this study was granted by the Ethics Committee at the University of Hull. 
Informed consent was obtained from all participants before their involvement in the research.

\paragraph{Consent for Publication}
Not applicable.

\paragraph{Data Availability}
The data that supports the findings of this study is not publicly available but may be made available to researchers upon reasonable request to the corresponding author.

\paragraph{Materials Availability}
Not applicable.

\paragraph{Code Availability}
The code that supports the findings of this study is not publicly available but may be made available to researchers upon reasonable request to the corresponding author.

\paragraph{Author Contributions}
S. Rose: conceptualisation, methodology, data collection and analysis, writing (original draft, review and editing). D. Chakraborty: supervision, writing (review and editing).

%\bigskip

%\bigskip\noindent
%Springer journals and proceedings: \url{https://www.springer.com/gp/editorial-policies}

%\bigskip\noindent
%Nature Portfolio journals: \url{https://www.nature.com/nature-research/editorial-policies}
%
%\bigskip\noindent
%\textit{Scientific Reports}: \url{https://www.nature.com/srep/journal-policies/editorial-policies}
%
%\bigskip\noindent
%BMC journals: \url{https://www.biomedcentral.com/getpublished/editorial-policies}
%\end{flushleft}

\bibliography{sn-bibliography}% common bib file

\end{document}